\pdfoutput=1

\documentclass[11pt]{article}

\usepackage{EMNLP2022}

\usepackage{times}
\usepackage{latexsym}

\usepackage{graphicx}
\usepackage{hyperref}
\usepackage{amsfonts}
\usepackage{amsmath}
\usepackage{tabularx}
\usepackage{booktabs}
\usepackage{array}
\usepackage{multirow}
\usepackage{subfigure}
\usepackage{arydshln}
\usepackage{xcolor}
\usepackage[normalem]{ulem}
\usepackage{float}
\newcommand{\ignore}[1]{}
\usepackage{mathtools}
\mathtoolsset{showonlyrefs,showmanualtags}
\expandafter\def\expandafter\UrlBreaks\expandafter{\UrlBreaks
      \do\a\do\b\do\c\do\d\do\e\do\f\do\g\do\h\do\i\do\j%
      \do\k\do\l\do\m\do\n\do\o\do\p\do\q\do\r\do\s\do\t%
      \do\u\do\v\do\w\do\x\do\y\do\z\do\A\do\B\do\C\do\D%
      \do\E\do\F\do\G\do\H\do\I\do\J\do\K\do\L\do\M\do\N%
      \do\O\do\P\do\Q\do\R\do\S\do\T\do\U\do\V\do\W\do\X%
      \do\Y\do\Z}

\newenvironment{itemizesquish}[2]{\begin{list}{\labelitemi}{\setlength{\itemsep}{#1}\setlength{\labelwidth}{#2}\setlength{\leftmargin}{\labelwidth}\addtolength{\leftmargin}{\labelsep}}}{\end{list}}

\usepackage[T1]{fontenc}

\usepackage[utf8]{inputenc}

\usepackage{microtype}

\usepackage{inconsolata}

\title{Abstractive Summarization Guided by \\ Latent Hierarchical Document Structure}

\newcommand{\emldisplay}[2]{\texttt{\href{mailto:#1}{#2}}}
\newcommand{\eml}[1]{\emldisplay{#1}{#1}}

\author{Yifu Qiu  \quad Shay B. Cohen \\
Institute for Language, Cognition and Computation\\
School of Informatics, University of Edinburgh \\
  10 Crichton Street, Edinburgh, EH8 9AB \\
  \medskip
  \eml{Y.QIU-20@sms.ed.ac.uk}, 
  \eml{scohen@inf.ed.ac.uk} \\}

\begin{document}
\maketitle
\begin{abstract}
Sequential abstractive neural summarizers often do not use the underlying structure in the input article or dependencies between the input sentences. This structure is essential to integrate and consolidate information from different parts of the text. To address this shortcoming, we propose a hierarchy-aware graph neural network (HierGNN) which captures such dependencies through three main steps: 1) learning a hierarchical document structure through a latent structure tree learned by a \emph{sparse} matrix-tree computation; 2) propagating sentence information over this structure using a novel message-passing node propagation mechanism to identify salient information; 3) using graph-level attention to concentrate the decoder on salient information. Experiments confirm HierGNN improves strong sequence models such as BART, with a 0.55 and 0.75 margin in average ROUGE-1/2/L for CNN/DM and XSum. Further human evaluation demonstrates that summaries produced by our model are more relevant and less redundant than the baselines, into which HierGNN is incorporated. We also find HierGNN synthesizes summaries by fusing multiple source sentences more, rather than compressing a single source sentence, and that it processes long inputs more effectively.\footnote{Code is available at \href{https://github.com/yfqiu-nlp/hiergnn}{https://github.com/yfqiu-nlp/hiergnn}.}

\end{abstract}

\section{Introduction}

\begin{table}[t]
\resizebox{\linewidth}{!}{%
\begin{tabular}{p{10cm}}
\toprule
\textbf{Article Sentences:}
\\ 
1. {\color[HTML]{00659a}The town is home to the prestigious Leander Club, which has trained more than 100 Olympic medal-winning rowers}. \\
\textit{- 2 sentences are abbreviated here.} \\
4. {\color[HTML]{9a0018}The Royal Mail has painted more than 50 postboxes gold following Team GB's gold medal haul at London 2012}. \\
5. Originally it said it was only painting them in  winners home towns, or towns with which they are closely associated. \\
6. Town mayor Elizabeth Hodgkin said: `` {\color[HTML]{00659a}We are the home of rowing} ... I feel very excited about it." \\
\textit{- 5 sentences are abbreviated here.} \\
12. The {\color[HTML]{006601}Henley-on-Thames postbox} was painted on Friday. \\
\textit{- one sentence is abbreviated here.} \\

\midrule
\textbf{Reference Summary:} 
{\color[HTML]{9a0018}The Royal Mail has painted a postbox gold} in the {\color[HTML]{006601}Oxford-shire town of Henley-on-Thames} - {\color[HTML]{00659a}in recognition of its medal} {\color[HTML]{00659a}winning rowing club}. \\

\midrule
\textbf{BART's Summary:} 
{\color[HTML]{006601}A postbox in Henley-on-Thames} has been {\color[HTML]{9a0018}painted gold as part of the Royal Mail's `` Olympic gold '' campaign}.
 \\

\midrule
\textbf{Our HierGNN's Summary:} 
{\color[HTML]{9a0018}A Royal Mail postbox} in {\color[HTML]{006601}Henley-on-Thames} has been {\color[HTML]{9a0018}painted gold} in {\color[HTML]{00659a}honour of the town 's Olympic rowing success}. \\
 
\bottomrule
\end{tabular}%
}
\caption{Example of an article from XSum with summaries given by human-written reference, BART \cite{lewis2020bart} and our HierGNN equipped with BART. BART's summary fails to capture all information pieces as the reference (as highlighted in various colors), while HierGNN has advantages in combining the information from multiple locations in the source side.}
\label{tab:summarization_illustration}
\end{table}

Sequential neural network architectures in their various forms have become the mainstay in abstractive summarization \cite{see-etal-2017-getTothePoint,lewis2020bart}. However, the quality of machine-produced summaries still lags far behind the quality of human summaries \cite{huang2020whatwehaveachievedinSummarization,xie-etal-2021-factual-consistency,cao-etal-2022-hallucinated-factuality,lebanoff2019scoringSentenceSingletons}. Due to their sequential nature, a challenge with neural summarizers is to capture hierarchical and inter-sentential dependencies in the summmarized document.

Progress in cognitive science suggests that humans construct and reason over a latent hierarchical structure of a document when reading the text in it \cite{graesser1994constructingInferenceDuringNarrativeTextComprehension,goldman1999narrative}. Such \textit{reasoning behavior} includes uncovering the salient contents and effectively aggregating all related clues spreading across the documents to understand the document. \citet{lebanoff2019scoringSentenceSingletons} found that human editors usually prefer writing a summary by fusing information from multiple article sentences and reorganizing the information in summaries (sentence fusion), rather than dropping non-essential elements in an original sentence such as prepositional phrases and adjectives (sentence compression). Different summarization benchmarks show there are between 60-85\% summary sentences that are generated by sentence fusing. These recent findings support our motivation to
make use of hierarchical document structure when summarizing a document.

We present a document hierarchy-aware graph neural network (HierGNN), a neural encoder with a reasoning functionality that can be effectively incorporated into any sequence-to-sequence (seq2seq) neural summarizer. 
Our HierGNN first learns a latent hierarchical graph 
via a sparse variant of the matrix-tree computation \cite{koo2007structuredpredictionMatrixTreeTheorm,liu-etal-2019-SummarizationasTreeInduction}. It then formulates sentence-level reasoning as a graph propagation problem via a novel message passing mechanism. During decoding, a graph-selection attention mechanism serves as a source sentence selector, hierarchically indicating the attention module which tokens in the input sentences to focus on.

Our experiments with HierGNN, incorporated into both pointer-generator networks \cite{see-etal-2017-getTothePoint} and BART \cite{lewis2020bart}, confirm that HierGNN substantially improves both the non-pretrained and pretrained seq2seq baselines in producing high-quality summaries. Specifically, our best HierGNN-BART achieves an average improvement of 0.55 and 0.75 points in ROUGE-1/2/L on CNN/DM and XSum.
Compared with a plain seq2seq model, HierGNN encourages the summarizers to favor sentence fusion more than sentence compression when generating summaries. Modeling the hierarchical document structure via our sparse matrix-tree computation also enables HierGNN to treat long sequences more effectively. In addition, our sparse adaptive variant of the matrix-tree computation demonstrates a more powerful expressive ability over the original one \cite{koo2007structuredpredictionMatrixTreeTheorm,liu-etal-2019-SummarizationasTreeInduction}.
We summarize our contributions as follows,

\begin{itemizesquish}{-0.3em}{0.5em}
    \item We present a novel encoder architecture for improving seq2seq summarizers. This architecture captures the hierarchical document structure via an adaptive sparse matrix-tree computation, with a new propagation rule for achieving inter-sentence reasoning.
    \item We design a graph-selection attention mechanism to fully leverage the learned structural information during decoding in advantages over only using it in encoding.
    \item Results on CNN/DM and XSum demonstrates the effectiveness of HierGNN in improving the quality of summaries for both non-pretrained and pretrained baselines. An in-depth analysis confirms our module improves the integration of information from multiple sites in the input article and that it is more effective in processing long sequence inputs.
\end{itemizesquish}

\section{Related Work}

\textbf{Neural Abstractive Summarization} \newcite{rush-etal-2015-neuralSummarization} first proposed to use a sequence-to-sequence model with an attention mechanism to perform sentence compression. \newcite{mendes-etal-2019-jointly} demonstrated the advantages and limitations of neural methods based on sentence compression. The pointer-generator networks (PGN; \citealt{see-etal-2017-getTothePoint}) enhances the attention model with a copying functionality. PGN has also been further extended to create summarization systems by incorporating the topic information \cite{topicAwarePGN}, document structural information \cite{song2018structureInfusedCopy}, semantic information \cite{hardy-vlachos-2018-AMRguidedSummarization}, and was improved by replacing the plain LSTM module with the more advanced Transformer model to overcome the difficulty in modeling long sequence input \cite{pilault-etal-2020-extractiveAbsTransformer,wang2021exploringExplainableSelectionControlAbsSummarization,fonseca2022}. For the pretrained models, BERTSum \cite{liu2019BERTSum} adopted the BERT encoder for the summarizer, with a randomly initialized decoder. \newcite{lewis2020bart} presented BART which pre-trains both the underlying encoder and decoder. \newcite{dou-etal-2021-gsum} investigated ``guidance signals'' (e.g., keywords, salient sentences) for further boosting the performances.

\noindent \textbf{Graph Neural Approach for Summarization} Graph neural networks have demonstrated their ability to capture rich dependencies in documents to be summarized. \newcite{wang2020heterogeneousGraphSum} use a ``heterogeneous graph'' with sentence nodes and co-occurring word nodes to capture the sentence dependencies. \newcite{jin2020semsum} use two separate encoders to encode the input sequence with a parsed dependency graph. 
\newcite{cui-etal-2020-enhancing-extsum-with-topic-gnn} use a bipartite graph with a topic model to better capture the inter-sentence relationships.
\newcite{kwon-etal-2021-considering-tree-structure-in-sent-extsummarization} capture both intra- and inter-sentence relationships via a nested tree structure.
\newcite{zhu2021enhancingFactualbyKG} use entity-relation information from the knowledge graph to increase the factual consistency in summaries.

Our approach is related to the structural attention model \cite{balachandran-etal-2021-structsum,liu-etal-2019-SummarizationasTreeInduction}, but differs in two major ways: (i) we introduce an adaptive sparse matrix-tree construction to learn a latent hierarchical graph and a novel propagation rule; (ii) we investigate to use the structure information both with the encoder and the decoder for abstractive summarization, and not just the encoder. These shows to be more effective for unsupervised learning of the latent hierarchical structure while can defeat the approach that leverages external graph constructor \cite{balachandran-etal-2021-structsum}.

\section{Hierarchy-aware Graph Neural Encoder}\label{sec:graph_construct}

HierGNN learns the document structure in an end-to-end fashion without any direct structure supervision, and does not need an external parser to construct the structure, unlike previous work \citep{balachandran-etal-2021-structsum,huang-etal-2020-knowledgegraphaugmentedsummarization,wang2020heterogeneousGraphSum,cardenas2022trade}. In addition, it empirically improves over supervised graph construction, which has been a challenge \cite{balachandran-etal-2021-structsum}. 

Sequential summarizers encode an $N$-token article, $X = (x_1, \cdots, x_N)$ as $d$-dimensional latent vectors using an encoding function $\mathbf{h}_{enc}(x_t) \in \mathbb{R}^{d}$ and then decodes them into the target summary $Y$. (We denote by $\mathbf{h}_{enc}(X)$ the sequence of $x_t$ encodings for $t \le N$.) Our model includes four modules in addition to this architecture: 1) a sparse matrix-tree computation for inferring the document hierarchical structure, ii) a novel message-passing layer to identify inter-sentence dependencies, iii) a reasoning fusion layer aggregating the outputs of the message-passing module; and vi) a graph-selection attention module to leverage the encoded structural information. 

\subsection{Learning the Latent Hierarchical Structure}

We first introduce our latent structure learning algorithm that makes use of a sparse variant of the matrix-tree theorem \cite{1986GraphTB,koo2007structuredpredictionMatrixTreeTheorm}. 

\noindent \textbf{Latent Document Hierarchical Graph.} 
We represent the document as a complete weighted graph, with each node
representing a sentence. The edge weights are defined as the marginal
probability of a directional dependency between two sentences.
 In addition, each sentence node has an extra probability value, the ``root probability'' which indicates  the \textit{hierarchical role} of the sentence, such as the roles of \emph{the lead}, \emph{most important facts}, or \emph{other information} defined based on the inverted pyramid model for news articles \cite{po2003news,ytreberg2001moving}. Intuitively, a sentence with a high root probability (high hierarchical position) conveys more general information; namely, it is a \textit{connector}, while a sentence with a lower root probability (\textit{information node}) carries details supporting its higher connectors. The underlying graph structure is latent and not fixed, summed out in our overall probability model using the matrix-tree theorem.

\begin{figure*}[]
    \centering
    \includegraphics[width=0.75\textwidth]{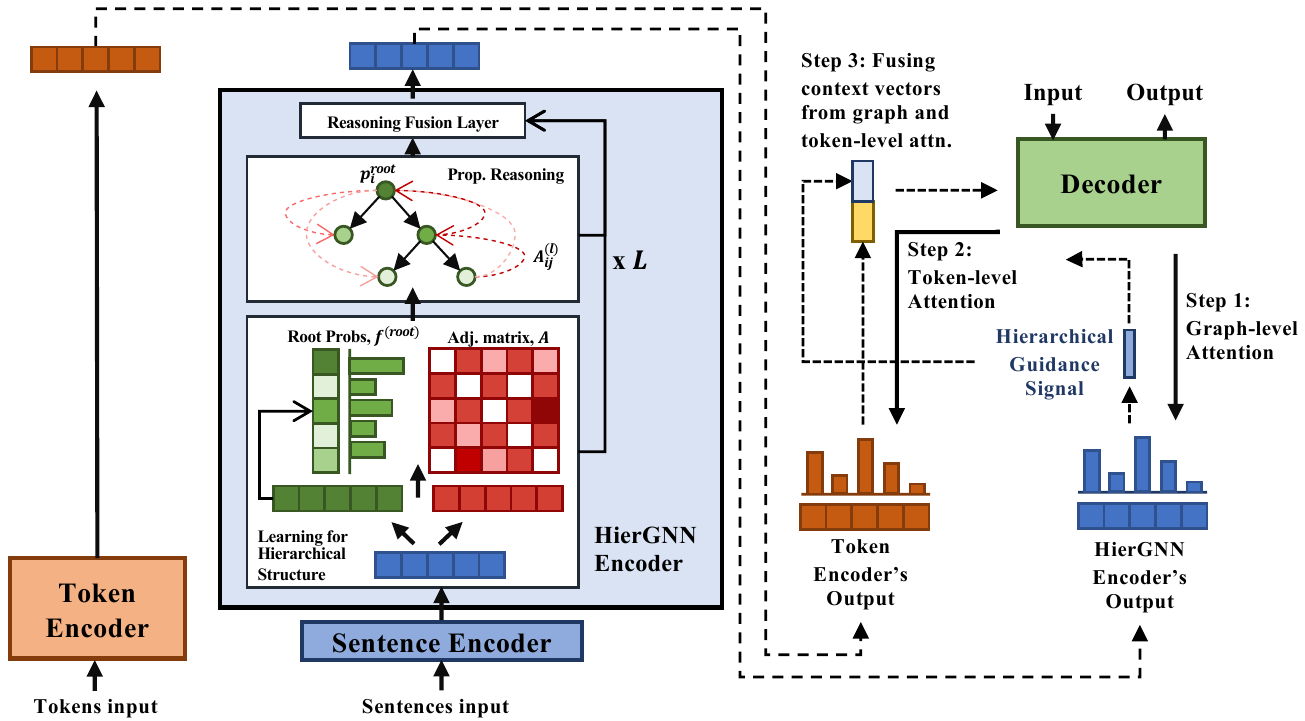}
    \caption{Architecture for the sequence-to-sequence model with HierGNN reasoning encoder.}
    \label{fig:hiergnn_layer}
\end{figure*}

\noindent \textbf{Sparse Matrix-Tree Computation.} For an article with $M$ sentences, we start from the sentence embeddings as the node initialization $H^{(0)} = [\mathbf{s}_1, ..., \mathbf{s}_i, ..., \mathbf{s}_M]$. We then use two independent non-linear transformations to obtain a pair of \textit{parent} and \textit{child} representation for each sentence,
\begin{align}
	\mathbf{s}_i^{(p)} &= \sigma(W_p\mathbf{{s}}_i+b_p), \\ \mathbf{s}_i^{(c)} &= \sigma(W_c\mathbf{{s}}_i+b_c),
\end{align}
\noindent where $W_p, W_c, b_p, b_c$ are parameters, $\sigma$ is the ReLU activation function \cite{dahl2013ReLU}.

The standard use of the matrix-tree theorem \cite{1986GraphTB} computation (MTC; \citealt{smith2007probabilistic,koo2007structuredpredictionMatrixTreeTheorm,mcdonald2007complexity}) includes the exponential function to calculate a matrix $F\in \mathbb{R}^{M\times M}$ with positive values with each element $f_{ij}$ representing the weight of the directional edge from a node $s_i$ to $s_j$; and a positive vector of root scores $\mathbf{f}^{(root)} \in \mathbb{R}^M$. However, having a dense matrix degrades our graph reasoning module by including irrelevant information from redundant $M$ sentence nodes. Inspired by the work about sparse self-attention \cite{zhang2021sparseReluAttention,correia2019adaptivelySparseTransformer}, we introduce an adaptive solution to inject sparsity into MTC. We replace the exponential scoring function with the ReLU function ($\mathrm{ReLU}(x \in \mathbb{R}) = \max \{ x, 0 \}$ and similarly coordinate-wise when $x$ is a vector) and calculate the root $f_i^{(root)}$ and edge scores $f_{ij}$ by a fully-connected layer and a bi-linear attention layer, respectively,
\begin{align}
    f_i^{(root)} &= \textsc{ReLU}(W_r \mathbf{s}_i^{(p)} + b_r) + \varepsilon, \\
    f_{ij} &= \textsc{ReLU}({{\mathbf{s}_i^{(p)}}^\top W_{bi}{\mathbf{s}_j^{(c)}} }) + \varepsilon,
\end{align}
\noindent where $W_{bi}, W_r, b_r$ are learnable. (We use $\varepsilon=10^{-6}$ to avoid matrix non-invertibility issues.) Compared to the exponential function, ReLU relaxes $F$ and $\mathbf{f}^{(root)}$ to be non-negative, thus being capable of assigning zero probability and pruning dependency edges and roots. We finally plug in these quantities to the standard MTC \cite{1986GraphTB} and marginalize the edge and root probabilities as the adjacency matrix $A(i,j)=P(z_{ij}=1)$ and root probability $p^{r}_{i}$ representing the hierarchical role (i.e., the likelihood to be a connector) of each sentence.


\subsection{Reasoning by Hierarchy-aware Message Passing} 

We present a novel message-passing mechanism over the learned hierarchical graph. This mechanism realizes the inter-sentence reasoning where connectors can aggregate information from their related information nodes while propagating the information to others. For the $i$-th sentence node, the edge marginal controls the aggregation from its $K$ information nodes; and the root probability controls the neighbouring information is combined as $i$-th node's update $\mathbf{u}^{(l)}$ in the $l$-th reasoning layer,
\begin{equation}
    \mathbf{u}^{(l)}_i = (1-p^{r}_{i}) \mathcal{F}_r(\mathbf{s}_i^{(l)}) + (p^{r}_{i}) \sum_{k=1}^K A_{ik} \mathcal{F}_n(\mathbf{s}_k^{(l)}),
\end{equation}
where $\mathcal{F}_r$ and $\mathcal{F}_n$ are parametric functions. Intuitively, if a sentence is a \textit{connector}, it should have strong connectivity with the related \textit{information nodes}, and aggregate more details. Each information node learns to either keep the uniqueness of its information or fuse the information from the connectors. To filter out the unnecessary information, we adopt a gated mechanism as the information gatekeeper in the node update,
\begin{align}
    \mathbf{g}_i^{(l)} &= \sigma (\mathcal{F}_g([\mathbf{u}_i^{(l)}; \mathbf{h}_i^{(l)}])), \\
    \mathbf{h}_i^{(l+1)} &= \text{LN}(\mathbf{g}_i^{(l)} \odot \mathcal{\phi}(\mathbf{u}_i^{(l)}) + (\mathbf{1}-\mathbf{g}_i^{(l)}) \odot \mathbf{h}_i^{(l)}),
\end{align}
where $\mathcal{F}_g$ is a parametric function and $\odot$ is the element-wise dot product. We use layer normalization (\textsc{LN}) to stabilize the output for the update function. The function $\sigma$ is the sigmoid function, and $\phi$ can be any non-linear function.

\subsection{Reasoning Fusion Layer} 
We construct \emph{reasoning chains} that consist of $L$ hops by stacking $L$ HierGNN blocks together. To handle cases where fewer than $L$ hops are needed, we add a fusion layer to aggregate the output from each reasoning hop to produce the final output of HierGNN. A residual connection is also introduced to pass the node initialization directly to the output,
\begin{equation}
    \mathbf{h}^{(G)}_i = (W_g[\mathbf{h}^{(1)}_i, ..., \mathbf{h}^{(L)}_i] + b_g) + \mathbf{h}^{(0)}_i,
\end{equation}
\noindent where $W_g, b_g$ are learnabale parameters. 
We use two approaches for layer use: (a) \textit{Layer-Shared Reasoning (LSR)}: we construct a shared reasoning graph first, followed by $L$ message passing layers for reasoning; (b) \textit{Layer-Independent Reasoning (LIR)}: we learn the layer-wise latent hierarchical graphs independently, where each message passing layer uses its own graph.

\subsection{Graph-selection Attention Mechanism}

In addition to token-level decoding attention, we propose a \textit{graph-selection attention mechanism} (GSA) to inform the decoder with learned hierarchical information, while realizing the sentence-level content selection. In each decoding step $t$, our decoder first obtains a graph context vector, $\mathbf{c}_G^t$, which entails the global information of the latent hierarchical graph. We first compute the graph-level attention distribution $\mathbf{a}_G^t$ by, 
\begin{align}
    e^t_{v_i} &= \textsc{Attn}^{(G)}(\mathbf{h}^{(L)},\mathbf{z}_t), \\ 
    \mathbf{a}_G^t &= \textsc{Softmax}(\mathbf{e}^t),
\end{align}
where $\textsc{Attn}^{(G)}$ is a graph attention function. The vectors $\mathbf{h}_i^{(L)} \in \mathbb{R}^d, \mathbf{z}_t \in \mathbb{R}^d$ are the $L$-th layer node embeddings for sentence $i$ and decoding state at time $t$, respectively. The graph context vector $\mathbf{c}_G^t \in \mathbb{R}^d$ is finally obtained by summing all $\mathbf{h}_i^{(L)}$ weighted by $\mathbf{a}_G^t$. The value of $\mathbf{c}_G^t$ is used as an additional input for computing token-level attention,
\begin{align}
     e_{i}^t &= \textsc{Attn}^{(T)}(\mathbf{h}_{enc}(X), \mathbf{z}_t,\mathbf{c}_G^t),
     \\
    \mathbf{a}_T^t &=  \textsc{Softmax}(\mathbf{e}^t),
\end{align}
where $\textsc{Attn}^{(T)}$ is a token-level attention function \cite{luong-etal-2015-effective,vaswani2017attention}. Again, the token-attentional context vector $\mathbf{c}_{f}^t$ is computed by summing the encoder outputs weighted by $\mathbf{a}_T^t$. The final context vector $\mathbf{c}_{f}^t$ is fused from the graph $\mathbf{c}_G^t$ and token context vectors $\mathbf{c}_T^t$ with a parametric function $g_{f}$, $\mathbf{c}_{f}^t = g_{f}(\mathbf{c}_G^t, \mathbf{c}_T^t)$.

\section{Experimental Setting}

\noindent\textbf{Benchmarks.} We evaluate our model on two common document summarization benchmarks. The first is the CNN/Daily Mail dataset \cite{hermann2015CNNDMdataset} in the news domain, with an average input of 45.7 sentences and 766.1 words, and a reference with an average length of 3.59 sentences and 58.2 words. We use the non-anonymized version
of \newcite{see-etal-2017-getTothePoint}, which has 287,084/13,367/11,490 instances for training, validation and testing. The second dataset we use is XSum \cite{narayan2018donXSum},
a more abstractive benchmark consisting of one-sentence human-written summaries for BBC news. The average lengths for input and reference are 23.26 sentences with 430.2 words and 1 sentence with 23.3 words, respectively. We follow the standard split of \newcite{narayan2018donXSum} for training, validation and testing (203,028/11,273/11,332).

\noindent \textbf{Implementations.} 
We experiment with the non-pretrained PGN of \newcite{see-etal-2017-getTothePoint} and the pretrained BART model \cite{lewis2020bart}. The implementation details are in Appendix~\ref{sec:model_implementations}.


\begin{table}[t]
\centering
\scalebox{0.75}{
\begin{tabular}{lcccccc}
\toprule
\multicolumn{1}{l}{\textbf{Non-pretrained}} & \textbf{R-1} & \textbf{R-2} & \textbf{R-L} & \textbf{BS}  \\ \toprule
LEAD-3 & 40.34    & 17.70            & 36.57      & -  \\
PGN                 & 39.53            & 17.28            & 36.38  & -     \\
StructSum ES  & 39.63            & 16.98            & 36.72      & -      \\
StructSum LS  & 39.52            & 16.94            & 36.71       & -   \\
StructSum (LS + ES)  & 39.62            & 17.00            & \textbf{36.95}        & 21.70    \\
\midrule
PGN - Ours                      & 39.07            & 16.97            & 35.87    & 23.74    \\
HierGNN-PGN (LSR)    & \textbf{39.87}	 & \textbf{17.77}	&  36.85 & \textbf{25.64}\\
HierGNN-PGN (LIR)     &  39.34	&   17.39	&   36.44 & 25.26   \\ 

\toprule
\multicolumn{1}{l}{\textbf{Pretrained}} & \textbf{R-1} & \textbf{R-2} & \textbf{R-L} & \textbf{BS}  \\ \toprule

BERTSUMABS         & 41.72       & 19.39       & 38.76    & 29.05    \\
BERTSUMEXTABS   & 42.13       & 19.60       & 39.18    & 28.72    \\
T5-Large         & 42.50       & 20.68       & 39.75    & -   \\
BART             & 44.16       & 21.28       & 40.90 & -    \\
Hie-BART             & 44.35       & 21.37       & 41.05 & -    \\
HAT-BART             & 44.48       & 21.31       & 41.52 & -    \\\midrule
BART - Ours     & 44.62	      & 21.49	    & 41.34 &  33.98    \\ 
BART + SentTrans.    & 44.44	      & 21.44	    & 41.27 &  33.90    \\ 
HierGNN-BART (LSR)   &   44.93  &	21.7    &	41.71  & 34.43  \\
HierGNN-BART (LIR)   &   \textbf{45.04} & 	\textbf{21.82} &	\textbf{41.82}   &  \textbf{34.59}  \\ \bottomrule
\end{tabular}}
\caption{Automatic evaluation results in ROUGE scores, BERTScore (BS) on CNN/DM
. The top and bottom blocks show the comparison for non-pre-training and pre-training models separately.
We use \textbf{bold} to mark the best abstractive model. 
}
\label{tab:cnndm_rouge}
\end{table}
\begin{table}[t]
\centering
\scalebox{0.78}{
\begin{tabular}{lcccc}
\toprule
\textbf{Non-pretrained}     & \textbf{R-1} & \textbf{R-2} & \textbf{R-L} & \textbf{BS}  \\ \midrule
LEAD-3    & 16.30             & 1.60              & 11.95            & -       \\
Seq2Seq (LSTM)  & 28.42            & 8.77             & 22.48            & -   \\
Pointer-Generator & 29.70             & 9.21             & 23.24  & 23.16    \\
PGN + Coverage  & 28.10             & 8.02             & 21.72            & -  \\ \midrule

HierGNN-PGN (LSR)    & 30.14	&   10.21	&   \textbf{24.32}    &       27.24   \\
HierGNN-PGN (LIR)     & \textbf{30.24}            & \textbf{10.43}             & {24.20}             &   \textbf{27.36}   \\

\toprule
\textbf{Pretrained}     & \textbf{R-1} & \textbf{R-2} & \textbf{R-L} & \textbf{BS}  \\ \midrule
BERTSUMABS      & 38.76       & 16.33       & 31.15  &  37.60   \\
BERTSUMEXTABS   & 38.81       & 16.50       & 31.27  &   38.14  \\
T5 (Large)        & 40.9        & 17.3        & 33.0    &  - \\
BART            & 45.14       & {22.27}       & {37.25}    &  - \\
HAT-BART            & \textbf{45.92}       & \textbf{22.79}       & \textbf{37.84}    &  - \\
\midrule
BART - Ours     & 44.97	&   21.68 & 	36.47 &  52.89    \\
BART + SentTrans.    & 45.12	      & 21.62	    & 36.46 &  52.95    \\ 
HierGNN-BART (LSR)   &   45.19	&   21.71	&   36.59   & 52.94  \\
HierGNN-BART (LIR)   &   {45.39}	&   {21.89}   &	{36.81}   &  \textbf{53.15}  \\
 \bottomrule

\end{tabular}}
\caption{Automatic evaluation results in ROUGE scores, BERTScore (BS) on XSum. All of our HierGNN-PGN models are trained without a coverage mechanism. 
We use \textbf{bold} for the best model.
}
\label{tab:xsum_result}
\end{table}

\noindent \textbf{Baselines.} We compare HierGNN with three types of baselines: 1) the base models for developing HierGNN; and 2) several strong non-pretrained and pretrained baselines; 3) abstractive summarizers boosted with the hierarchical information. 

We compare HierGNN-PGN with the non-pretrained baselines. We first include the \textbf{LEAD-3} \cite{nallapati2017summarunner} that simply selects the top three sentences in the article as the summary. \textbf{StructSum} \cite{balachandran-etal-2021-structsum} is a PGN-based model, which incorporates structure information by an explicit attention mechanism (ES Attn) on a coreference graph and implicit attention mechanism (IS Attn) on an end-to-end learned document structure. StructSum ES+IS Attn uses both implicit and explicit structures. 

We compare HierGNN-PGN with the pretrained baselines. \textbf{BERTSumAbs} and \textbf{BERTSumExtAbs} are two abstractive models by \newcite{liu2019BERTSum} based on the BERT encoder. We also incorporate a strong multitask sequence generation model, \textbf{T5-Large}. \textbf{Hie-BART} \cite{akiyama-etal-2021-hieBart} enhances BART by jointly modeling the sentence and token-level information in the self-attention layer. \textbf{HAT-BART} \cite{rohde2021hierarchicalBART} appends a sentential Transformer block on top of BART's encoder to model the sentence-level dependencies. We also develop a baseline, \textbf{BART+SentTrans.}, replacing our MTC block with a Transformer block. This baseline uses a comparable number of parameters to our HierGNN. We aim to verify the advantage of modeling the document's hierarchical information by MTC over just increasing the model size.

\section{Results}

\begin{table}[]
\centering
\scalebox{0.8}{
\begin{tabular}{ccccc}
\toprule
\textbf{Model}     & \textbf{Rel.}          & \textbf{Inf.}          & \textbf{Red.}          & \textbf{Overall}       \\ \midrule
BERTSUMABS   &  *-0.43        & *-0.33      & -0.11          & *-0.29          \\
T5           & 0.08          & -0.09        & 0.05         &    0.01     \\
BART         & 0.15          & \textbf{0.24}         & -0.04         & 0.12          \\
HierGNN-BART & \textbf{0.20} &  0.19        & \textbf{0.09} & \textbf{0.16} \\
\bottomrule
\end{tabular}}
\caption{Results for the human evaluation based on i) Relevance (Rel.), ii) Informativeness (Inf.), and iii) Redundancy (Red.). * indicates statistically significant improvements over the baselines with our model (*: by pair-wise t-test with $p < 0.05$, corrected using Benjamini–Hochberg method to control the False Discovery Rate \cite{benjamini1995controllingfalsedicoveryrate(fdr)} for multiple comparison). We \textbf{bold} the best results in each criteria and the overall evaluation. Detailed results are given in Appendix~\ref{sec:human_eval_appendix}.}
\label{tab:human_eval}
\end{table}

\noindent \textbf{Automatic Evaluation.} We evaluate the quality of summaries through ROUGE F-1 scores \cite{lin-och-2004-rougeL} by counting the unigram (R-1), bigram (R-2) and longest common subsequence (R-L) overlaps. To avoid the use of pure lexical overlap evaluation \cite{huang2020whatwehaveachievedinSummarization}, we also use BERTScore \cite{zhang2019bertscore}.

\begin{table}[]
\centering
\scalebox{0.8}{
\begin{tabular}{lcccc}
\toprule
\multicolumn{1}{c}{}      & \textbf{R-1} & \textbf{R-2} & \textbf{R-L} & \textbf{BS} \\ \midrule
\textbf{Full Model}                & 30.24	 & 10.43	& 24.20   &  27.36
             \\ \midrule
w/o HierGNN Module               & -0.54                      & -1.22                      & -0.96                     &      -4.20
                  \\
w/o Graph-select (GSA)           & -0.41                      & -0.41                      & -0.17                     &            -0.27
            \\
w/o Sparse MTC                    & -0.14                      & -0.25                      & +0.05                    &         -0.41               \\
w/o Graph Fusion                  &              -0.94	&   -0.81	&   -0.77             &    -1.39                \\
\bottomrule
\end{tabular}}
\caption{Ablation study of each modules in our HierGNN-PGN (LIR) model on XSum.}
\label{tab:ablation}
\end{table}

We summarize the results for non-pretrained and pretrained models on CNN/DM and XSum in the upper and bottom block of Table~\ref{tab:cnndm_rouge} and Table~\ref{tab:xsum_result}, respectively. Our HierGNN module improves the performance over the PGN and BART for both CNN/DM and XSum, demonstrating the effectiveness of our reasoning encoder for the non-pretrained and pretrained summarizers. Secondly, the best model of HierGNN-PGN achieves higher scores than StructSum ES and ES+IS that explicitly construct the document-level graph representation using an external parser in pre-processing. This indicates our learned hierarchical structure can be effective and beneficial for downstream summarization without any supervision. HierGNN-BART also outperforms Hie-BART, HAT-BART and BART+SentTrans., which indicates that the MTC encoder's inductive bias is effective in modeling useful structure.

\begin{table}[t]
\centering
\scalebox{0.70}{
\begin{tabular}{lrr}
\toprule
\textbf{Model} & \textbf{Coverage} ($\nearrow$)  & \textbf{Copy Length} ($\searrow$) \\ \midrule
Reference      & \textbf{20.27 $\%$}                     & \textbf{5.10 }                    \\ \midrule
Pointer-Generator         & 11.78 $\%$                     & 18.82                    \\
Ours $w/o$ Graph Select Attn.    & 13.74 $\%$                     & 18.88                    \\
Ours $w/$ Graph Select Attn.     & \textbf{15.22} $\%$           &         \textbf{16.80}                 \\
\bottomrule
\end{tabular}}
\caption{Results of average copying length of sequences and coverage of the source sentences for the CNN/DM datasets. Arrows ($\nearrow$ or $\searrow$) indicate that larger or lower scores are better, respectively. }
\label{tab:copy_coverage_analysis}
\end{table}

\noindent \textbf{Human Evaluations.} We also invited human referees from Amazon Mechanical Turk to assess our model and additional three pure abstractive baselines including BERTSUMABS, T5-Large, BART on CNN/DM testing set. Our assessment focuses on three criteria: i) Relevance (\textit{Whether the conveyed information in the candidate summary is relevant to the article}?), ii) Informativeness (\textit{How  accurate and faithful information does the candidate summary convey}?), and iii) Redundancy (\textit{Whether the sentences in each candidate summary are non-redundant with each other}?). The detailed settings for human evaluation are presented in Appendix~\ref{sec:human_eval_details}. We ask the referees to choose the best and worst summaries from the four candidates for each criterion. The overall scores in Table~\ref{tab:human_eval} are computed as the fraction of times a summary was chosen as best minus the fraction it was selected as worst. The results show that our HierGNN-BART achieves the overall best performance. Moreover, while BART has a slightly better informativeness score, HierGNN-BART produces better summaries in terms of Relevance and Redundancy.

\noindent \textbf{Ablations.} We conduct an ablation study (in Table \ref{tab:ablation}) of the HierGNN encoder, graph-selection attention, sparse MTC and graph fusion layer. The ablation is done on our HierGNN-PGN LIR model  trained on XSum. The ablation in HierGNN reasoning module significantly degrades the model, which suggests the positive contribution of the functionality in across-sentence reasoning. The scores without GSA also confirm the guidance of graph-level information is beneficial. By removing the graph fusion layer, we again observe the performance decreases, which proves the benefits of fusing the neighbor feature from multiple hopping distances. Finally, the results also confirm the superiority of the sparse MTC over the dense MTC for learning effective hierarchical structure for summarization. 

\begin{table}[t]
\centering
\scalebox{0.9}{
\begin{tabular}{lccc}
\toprule 
             & \textbf{R-1} & \textbf{R-2} & \textbf{BS} \\ \midrule
BART         & 49.41        & 21.70        & 19.12       \\
HierGNN-BART & \textbf{49.62}        & \textbf{21.74}        & \textbf{20.32}      \\
\bottomrule
\end{tabular}}
\caption{Summarization performance on PubMed. 
We test BART and HierGNN-BART with the same hyperparameters settings.}
\label{tab:pubmed-result}
\end{table}

\begin{figure}[t]
    \centering
    \includegraphics[width=0.9\linewidth]{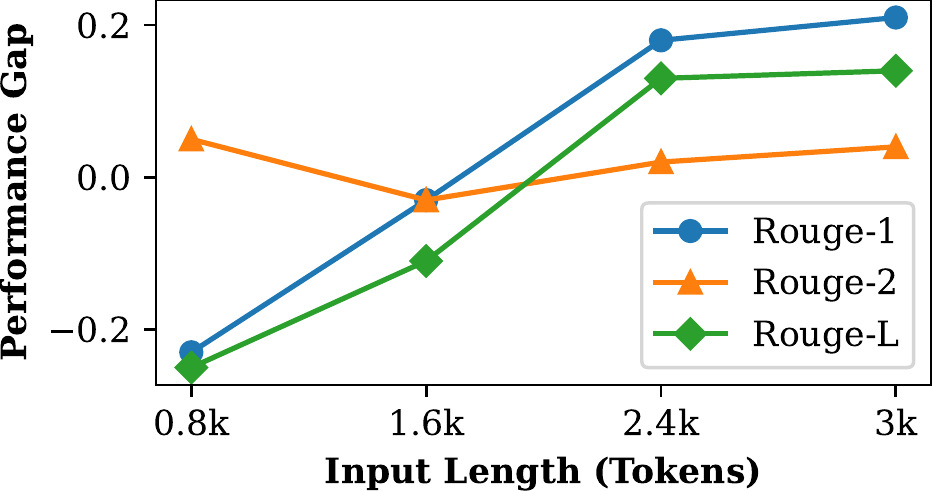}
    \caption{Performance gap on PubMed between HierGNN-BART with BART when summarizing articles truncated at different lengths. The gap between HierGNN and BART consistently increases with input length.}
    \label{fig:pubmed-res}
\end{figure}

\section{Discussion}

\noindent \textbf{Coverage and Copy Length.} We report two metrics introduced by \newcite{see-etal-2017-getTothePoint} in Table~\ref{tab:copy_coverage_analysis}. The coverage rate measures how much information in the source article is covered by the summary, while the average copy length indicates to what extent that summarizer directly copies tokens from the source article as its output. 
The higher coverage rate achieved by our HierGNN indicates that it can produce summaries with much richer information in the source article. 
\citeauthor{balachandran-etal-2021-structsum} find that PGN tends to over-copy content from the source article thus degenerating into an extractive model, particularly with more extractive datasets such as CNN/DM. We find that the graph-selection attention significantly reduces the average copy length, indicating that it informs the decoder to stop copying by leveraging the learned structural information in the encoder and that it reduces the reliance on PGN's copying functionality \cite{see-etal-2017-getTothePoint}. 
We show a qualitative example for the graph-selection attention outcome in Appendix \ref{sec:gsa_analysis}.

\begin{table}[]
\scalebox{0.79}{
\begin{tabular}{lcccc}
\toprule
\textbf{CNN/DM}    & \textbf{Comp.}          & \textbf{2-hop}          & \textbf{3-hop}          & \textbf{4-hop}         \\ \midrule
Reference & 63.03          & 32.08          & 4.59           & 0.31          \\ \midrule
BART      & 79.52          & 17.81          & {2.43}  & {0.24} \\
HierGNN-BART   & {78.13}($\downarrow$) & {19.29}($\uparrow$) & 2.36($\downarrow$)           & 0.21($\downarrow$)          \\ \midrule \midrule
\textbf{XSum}      & \textbf{Comp.}          & \textbf{2-hop}          & \textbf{3-hop}          & \textbf{4-hop}         \\ \midrule
Reference & 34.87          & 42.50           & 18.79          & 3.83          \\ \midrule
BART      & 28.47          & 42.51           & 23.05          & {5.98}          \\
HierGNN-BART   & {27.27}($\downarrow$) & {42.53}($\uparrow$) & {24.31}($\uparrow$) & 5.89($\downarrow$) \\ \bottomrule
\end{tabular}}
\caption{Percentages of summary sentences are synthesized by compression (information is extracted from a single source sentence) and fusion (information is combined from two or more source sentences). We use $\downarrow$ and $\uparrow$ to mark the changes between BART and HierGNN. }
\label{tab:fusion-analysis}
\end{table}

\noindent \textbf{Layer-shared or Layer-independent Reasoning?} In Tables \ref{tab:cnndm_rouge} and \ref{tab:xsum_result}, we observe that the layer-shared reasoning (LSR) architecture for HierGNN-PGN on CNN/DM outperforms the layer-independent reasoning (LIR) architecture, with the opposite being true for XSum. We attribute this difference to the inductive bias of the base model and the essential difference between the CNN/DM and XSum datasets. PGN-based models tend to copy and degenerate the model into an extractive summarizer \cite{balachandran-etal-2021-structsum}. With a more extractive dataset like CNN/DM, a complex reasoning procedure for the PGN-based model may not be necessary; instead, learning a single hierarchical structure and selecting the sentences to be copied accordingly is sufficient. However, XSum summaries are abstractive, and the dataset emphasizes combining information from multiple document sites (see discussion by \citealt{narayan2019article}). LIR then shows its advantage by learning separate hierarchical structure in each layer. For an abstractive base model (BART), LIR consistently outperforms LSR on both CNN/DM and XSum.

\noindent \textbf{Compression or Fusion?} To assess whether sentence fusion happens often, we quantify the ratio of sentence compression and sentence fusion that the model uses to generate summaries in Table~\ref{tab:fusion-analysis} \cite{lebanoff2019scoringSentenceSingletons}. In comparison to BART, HierGNN reduces the proportion of sentence compression in both CNN/DM and XSum. Furthermore, the summarization models tend to adopt sentence compression more than exists in human-written references for CNN/DM, while more sentence fusion is used for XSum. This observation reveals that mechanism learned by end-to-end for neural summarizers to produce summaries is different than that humans use. Human editors can flexibly switch between compression and fusion; the summarization models tend to adopt one of them to produce the output.

\begin{figure}
\captionsetup[subfigure]{labelformat=empty}
\centering
\begin{subfigure}{}
  \begin{minipage}[]{\linewidth}
  \includegraphics[width=1\linewidth]{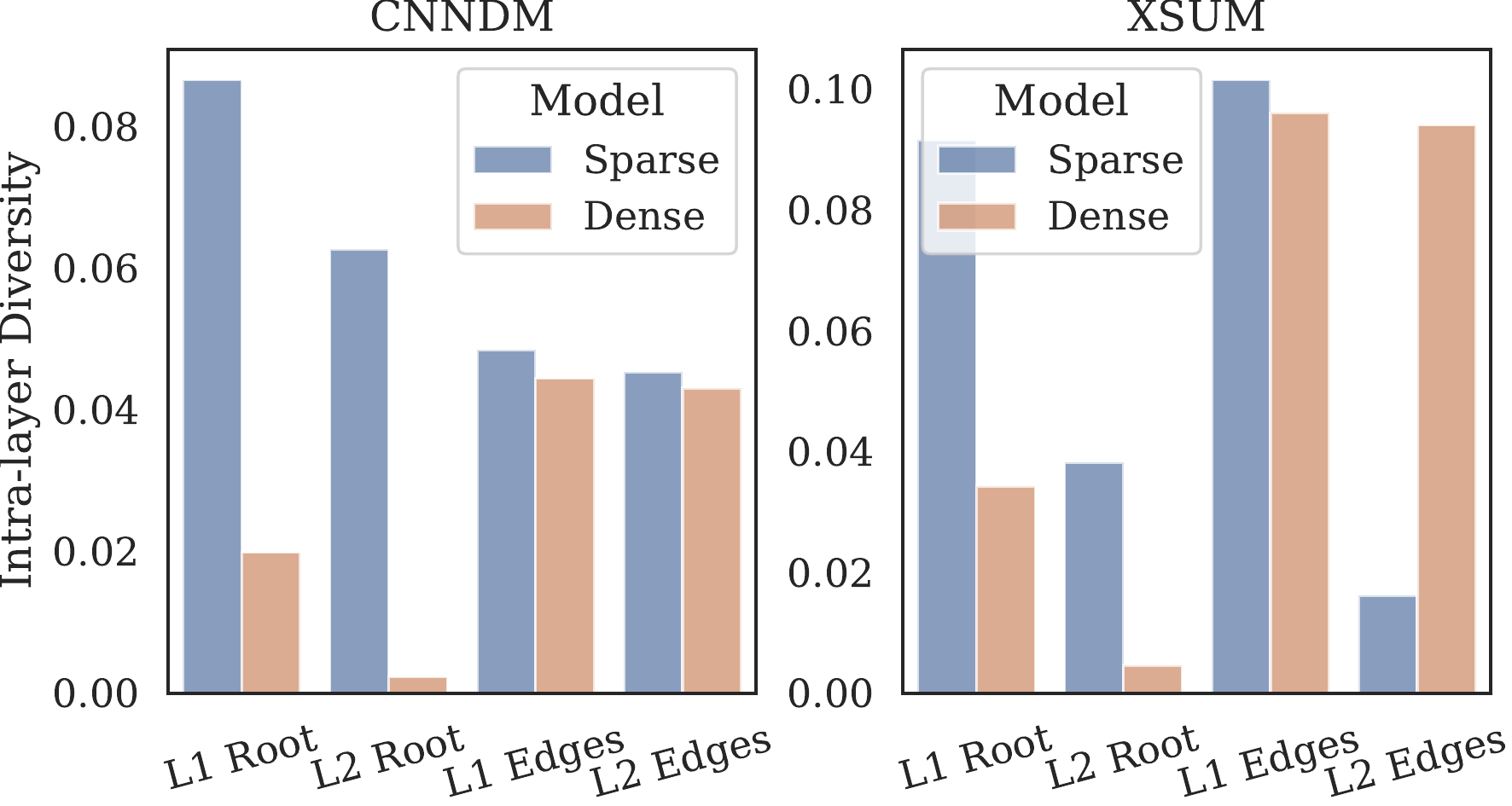} \\ \vspace{0.3cm}
  \includegraphics[width=1\linewidth]{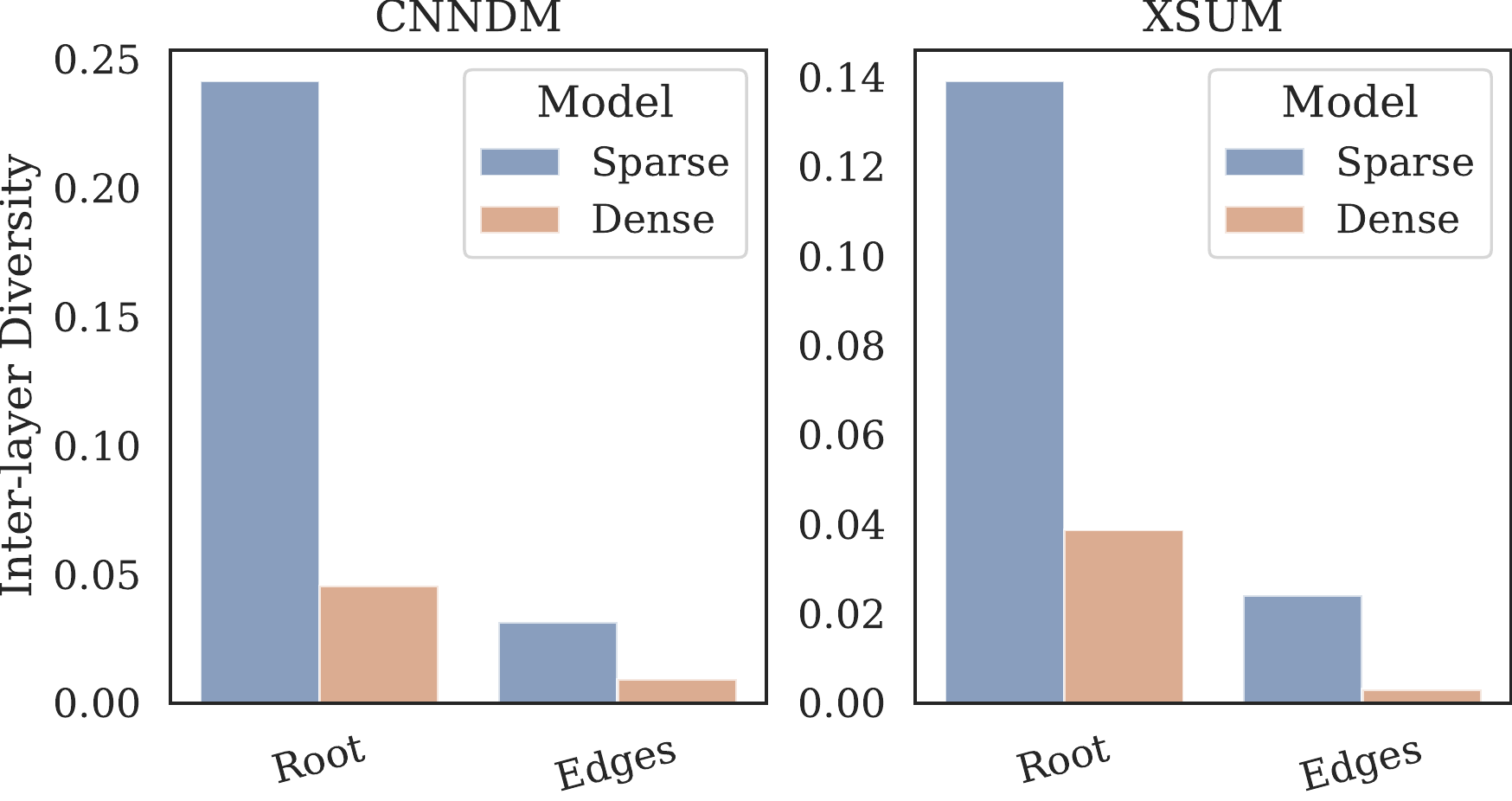}
  \end{minipage}
\end{subfigure}%
\vspace{-0.4cm}
\caption{Layer-wise intra-layer diversity (top) and inter-layer diversity (bottom) for BART with 2-layer HierGNN equipped with Sparse and Dense MTC.}
\label{fig:mtc-similarity-measure}
\end{figure}

\noindent \textbf{Effectiveness for Longer Sequence.} The performance of sequence-to-sequence models decays as the length of the input sequence increases \cite{j.2018generating-wikipedia-by-summarizaing-long-sequences} because they do not capture long-range dependencies. We hypothesize that HierGNN has a better capability in capturing such dependencies via its learned document hierarchical structure, thus enhancing the performance for long-sequence inputs. To verify this, we further conduct experiments on PubMed \cite{cohan2018ArxivPubMedBenchmark}, a long-document summarization dataset with scientific articles in the medical domain.
We summarize the performance in Table \ref{tab:pubmed-result}. We notice that HierGNN improves BART by a large margin. We further evaluate the advantages of HierGNN over vanilla BART with respect to inputs of various lengths. As shown in Figure~\ref{fig:pubmed-res}, when the input is longer than 1.6K tokens, HierGNN has a positive advantage over BART. As the input length increases, the advantage of HierGNN consistently becomes larger.

\begin{figure}[t]
    \centering
    \includegraphics[width=\linewidth]{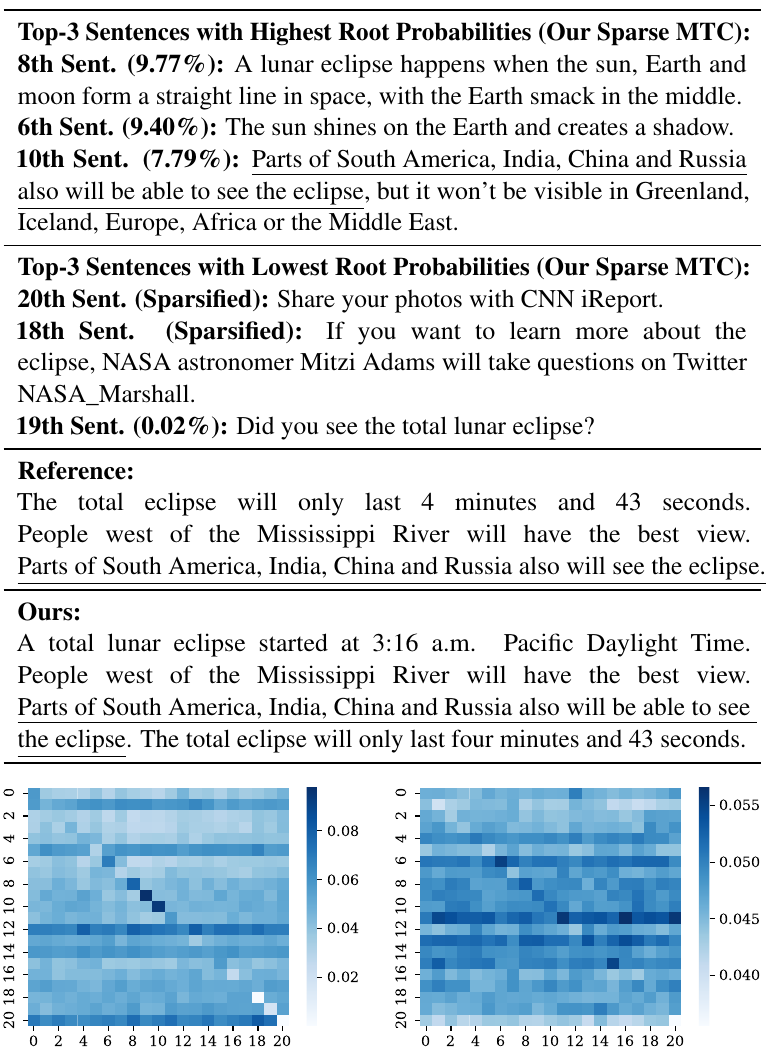}
    \caption{Top: the top-3 sentences with highest/lowest root probabilities, reference and summaries in article 23 in CNN/DM testing split. We underline the relevant contents; Bottom: visualizations for our sparse (Left) and the dense (Right) MTC layer for HierGNN-BART.}
    \label{fig:reasoning_case}
\end{figure}

\noindent \textbf{Sparse MTC or Dense MTC?} We also study the expressive ability of our adaptive sparse variant of the matrix tree computation. We design two quantitative metrics: 1) \textit{Intra-layer diversity} measures the diversity for the marginal distributions of roots and edges in each MTC layer, which is calculated by the range of the probability distribution; 2) \textit{Inter-layer diversity} measures the diversity for the marginal distributions of roots and edges between MTC layers, which is calculated by the average Jensen-Shannon (JS) Divergence between the marginal distributions of roots and edges in different layers \cite{zhang2021sparseReluAttention,correia2019adaptivelySparseTransformer}. We compare both intra-layer and inter-layer diversity for our adaptively sparse MTC and the original dense MTC \cite{koo2007structuredpredictionMatrixTreeTheorm,liu-etal-2019-SummarizationasTreeInduction,balachandran-etal-2021-structsum}. 


Figure~\ref{fig:mtc-similarity-measure} shows that our sparse variant of MTC has a higher diversity in both intra- (Top) and inter-layer (Bottom) metrics for CNN/DM and XSum, indicating that our sparse MTC has a more powerful expressive ability than dense MTC. We find that the sparsity of HierGNN is different across layers and datasets: 1) 99.66\% of HierGNN's predictions for XSum instances have at least one element that is sparsified to zero, while this proportion is 24.22\% for CNN/DM; 2) Almost all the sparsified elements in HierGNN's predictions for XSum are edges, while roots for CNN/DM; 3) 90.32\% of the elements of the edge distribution in the second MTC layer are sparsified in XSum, but no any sparsified element in the first layer. In CNN/DM, the proportion of sparsified elements in the first and second layer are almost identical. These observations reveal that sparse MTC can adaptively choose whether sparse out elements in root or edge distributions, thus boosting the richness of the structural information represented by MTC.

We finally show a qualitative case with three sentences per article, having the highest or lowest root probabilities (see Figure~\ref{fig:reasoning_case}), and the heatmap visualization of the learned hierarchical structures from sparse and dense MTC. We observe that the highest-probability root sentences tend to be summary-worthy while also scattering in different positions of the article, and the lowest probability is irrelevant. The structure learned by Sparse MTC tends to be more diverse and can successfully sparsify out the sentence nodes with irrelevant contents, e.g., 18th and 20th sentence.

\section{Conclusion}

We propose HierGNN that can be used in tandem with existing generation models. The module learns the document hierarchical structure while being able to integrate information from different parts of the text as a form of reasoning. Our experiments verify that HierGNN is effective in improving the plain sequential summarization models. 

\section*{Limitations}

The inductive bias of our HierGNN model has an assumption that the source article follows an ``inverted pyramid'' style of writing. This may pose limitations in the generalization of our model to other categories of input documents with no or a weak hierarchical structure. Future work includes understanding the limitations of HierGNN in different input domains (e.g., conversation summarization). Additionally, as other large-scale pretrained neural summarizers, our approach with an additional HierGNN encoder increases model complexity. To train our BART-based system, GPUs with at least 32GB of memory are required. Future work may focus on distilling the large HierGNN model into a much smaller size while retaining its original performance.

\section*{Ethical and Other Considerations}

\paragraph{Human evaluations.}
Human workers were informed of the intended use of the provided assessments of summary quality and complied with the terms and conditions of the experiment, as specified by Amazon Mechanical Turk.\footnote{\url{https://www.mturk.com}}
In regards to payment, workers were compensated fairly with the wage of \pounds9 hourly (higher than the maximum minimum wage in the United Kingdom) i.e.\ \pounds4.50 \, per HIT at 2 HITs per hour.\footnote{\url{https://www.gov.uk/national-minimum-wage-rates}}

\paragraph{Computing time.}

We first report the computing time for our most computationally intense HierGNN-BART (471 million parameters) using NVIDIA Tesla A100 with 40G RAM: with CNN/DM, the training takes around 81 GPU hours, and the inference takes 9.39 GPU hours. With XSum, the training takes around 32 GPU hours, and the inference takes 4.41 GPU hours.

Additionally, training of HierGNN-PGN (32 million parameters) on CNN/DM takes 0.79 seconds per iteration using 1 NVIDIA V100 GPU card with 16GB. We estimate the inference time is 4.02 documents per second.

\appendix

\section*{Acknowledgements}

We thank Zheng Zhao, Marcio Fonseca and the anonymous reviewers for their valuable comments.
The human evaluation was funded by a grant from the Scottish Informatics and Computer Science Alliance (SICSA). This work was supported by computational resources provided by the EPCC Cirrus service (University of Edinburgh) and the Baskerville service (University of Birmingham).


\bibliographystyle{acl_natbib}
\bibliography{custom}

\newpage
\appendix

\section{Implementation Details}
\label{sec:model_implementations}
\textbf{HierGNN-PGN} is developed based on the Pointer-Generator Network \cite{see-etal-2017-getTothePoint}.\footnote{\url{https://github.com/atulkum/pointer_summarizer}}
 To obtain the sentence representations, we use a CNN-LSTM encoder to capture both the $n$-gram features and sequential features \cite{kim-2014-CNNTextClassification,zhou2015c-lstmTextClassification}. The CNN's filter
windows sizes are set to be $\{1, 2, 3, 4, 5, 7, 9\}$ with 50 feature maps each.
 We set the dimension of the representations to be 512. The number of reasoning layers $L$ is set to 3 after a development set search in $\{1, 2, 3, 5, 10\}$. Other settings follow the best hyperparameters for CNN/DM as in \cite{see-etal-2017-getTothePoint}, and we use 60K iterations to train the coverage mechanism. For XSum, we discard the coverage training due to its redundancy for extreme summarization \cite{narayan2018donXSum}, and we use a beam of size 6. We search the best model by the validation ROUGE scores on both datasets with one search trial per hyperparameter.

\begin{table}[h]
\centering
\scalebox{0.75}{
\begin{tabular}{ccccc}
\toprule
\textbf{\#Layer} & \textbf{Val. PPL} ($\searrow$) & \textbf{R-1} ($\nearrow$)   & \textbf{R-2} ($\nearrow$)   & \textbf{R-L} ($\nearrow$)  \\ \midrule
1       & 8.61     & 30.06 & 10.09 & 24.23 \\
2       & 8.58     & 29.94 & 10.00 & 24.13 \\
\textbf{3}       & \textbf{8.51}     & \textbf{30.24} & \textbf{10.43} & 24.20 \\
5       & 8.54     & 30.14 & 10.23 & \textbf{24.32} \\
10      & 8.61     & 29.99 & 9.93  & 24.13 \\ \bottomrule
\end{tabular}}
\caption{Performance of HierGNN-PGN (LIR) on XSum with respect to the number of reasoning layers. ($\nearrow$) and ($\searrow$) indicates the larger and lower is better, respectively.}
\label{tab:layer_comp}
\end{table}

\noindent \textbf{HierGNN-BART} uses the pretrained architecture BART \cite{lewis2020bart}.\footnote{\url{https://github.com/facebookresearch/fairseq/tree/main/examples/bart}} We use the same approach to obtain the sentence representation as in \cite{akiyama-etal-2021-hieBart}. On top of the sentence encoder, we add a two-layer HierGNN to boost the sentence representations. The GSA for HierGNN-BART is implemented as the cross-attention in Transformer decoder, which first attends to the output of the reasoning encoder then the token encoder. For both CNN/DM and XSum, we follow the same fine-tuning settings as in \cite{lewis2020bart} except that we use 40K and 20K training steps for each dataset. We search the best model by the label smoothed cross entropy loss on validation set with one search trial per hyperparameter.

\noindent \textbf{Evaluation Metrics.} 
We use the implementation for ROUGE \cite{lin-och-2004-rougeL} from Google Research.\footnote{\url{https://github.com/google-research/google-research/tree/master/rouge}} We use the official implementation\footnote{\url{https://github.com/Tiiiger/bert_score}} for BERTScore \cite{zhang2019bertscore}. BERTScore is used with model setting in \texttt{roberta-large\_L17\_no\-idf\_version=0.3.9} as suggested.  

\noindent \textbf{Datasets.} We describe all our pre-processings for the used datasets as followed,
\begin{itemize}
    \item \textbf{CNN/DM}: For HierGNN-PGN, we directly use the data processed by  \citeauthor{see-etal-2017-getTothePoint}.\footnote{\url{https://github.com/abisee/pointer-generator}} For HierGNN-BART, we remain all the pre-processing steps to be the same as \citeauthor{lewis2020bart}.\footnote{\url{https://github.com/artmatsak/cnn-dailymail}}
    \item \textbf{XSum}: Following \citeauthor{lewis2020bart}, we do not pre-process the XSum dataset, and use the original version in \cite{narayan2018donXSum}.\footnote{\url{https://github.com/EdinburghNLP/XSum}}
    \item \textbf{PubMed}: We use the same pre-processing script in \url{https://github.com/HHousen/ArXiv-PubMed-Sum}. We remove the instances with article have less 3 sentences or abstract have less 2 sentences. We also remove three special tokens: newlines, \texttt{<S>} and \texttt{</S>}.
\end{itemize}

\begin{figure*}[t!]
    \centering
    \includegraphics[width=1.0\textwidth]{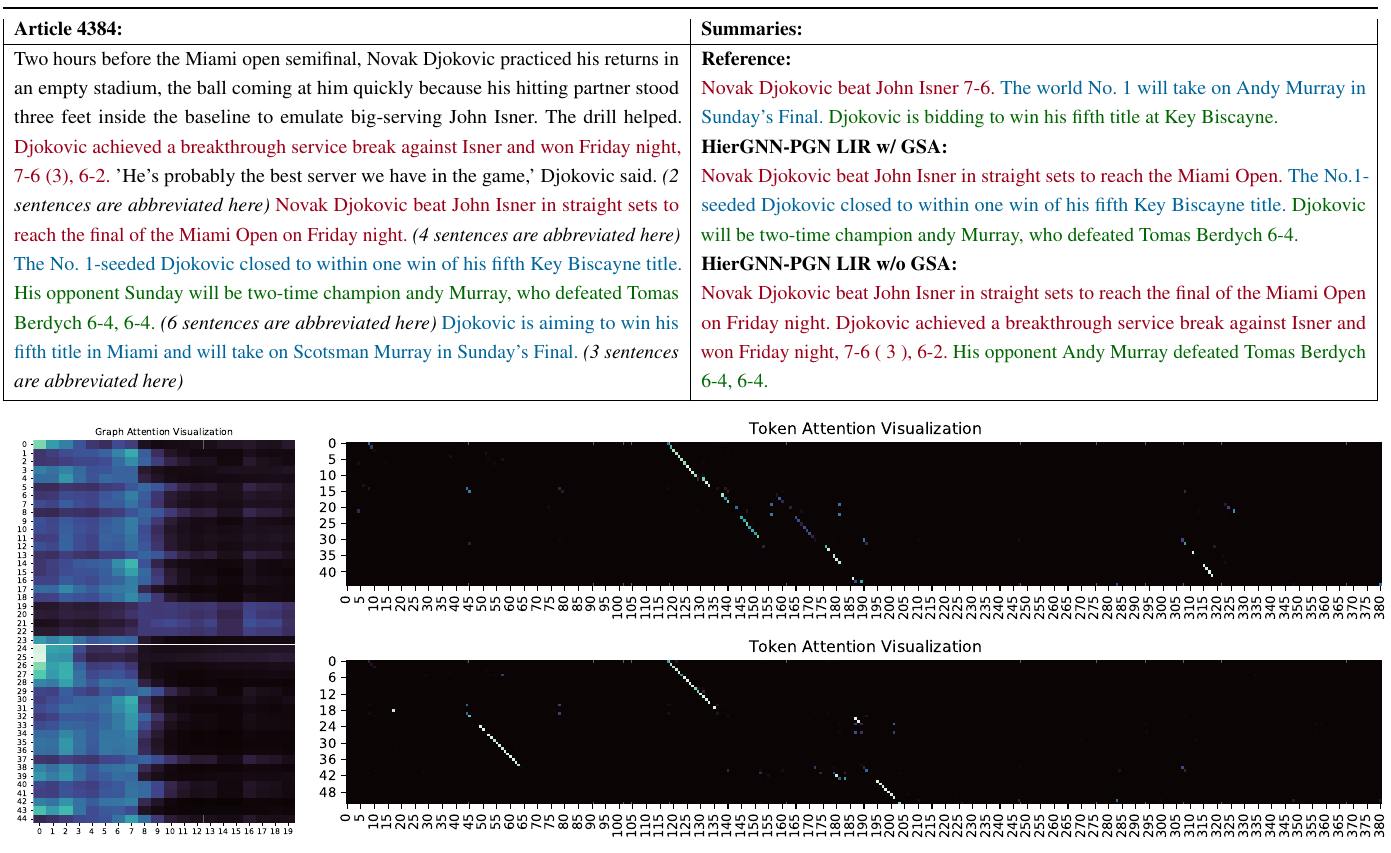}
    \caption{Top Table: CNN/DM testing article 4384 and produced summaries; Bottom Figure: visualization for GSA (left) and HierGNN LIR's token-level attention w/ GSA (right-bottom), and HierGNN-PGN LIR w/o GSA (right-top). X-axis, Y-axis are the encoding and decoding steps, respectively.}
    \label{fig:gsa_case_study}
\end{figure*}

\section{Details for Human Evaluation}\label{sec:human_eval_details}
We adopt several settings to control the quality of human evaluation: 1) we only use data instances whose length difference between candidate summaries does not exceed 35 tokens \cite{sun-etal-2019-compareSummarizerwoTargetLength,wu-etal-2021-bass}. 2) When publishing the tasks on MTurk, we require all referees to be professional English speakers located in one of the following countries: i) Australia, ii) Canada, iii) Ireland, iv) New Zealand, v) the United Kingdom and vi) the United States, with the HIT Approval Rate and number of HITs Approved to be greater than 98\% and 1,000. 3) We evaluate 25 instances in CNN/DM testing set in total, while each task is evaluated by three workers on MTurk. These settings give us the results with an inter agreement in the average of 58.96\%, 64.92\% and 51.52\% for Relevance, Informativeness and Redundancy, separately.

\section{Detailed Results for Human Evaluation}\label{sec:human_eval_appendix}

We show the detailed proportions for each choice in human evaluation in Table~\ref{tab:detalied_human}.
\begin{table}[h!]
\scalebox{0.85}{
\begin{tabular}[width=0.8\linewidth]{cccc}
\toprule
\textbf{Rel.}       & \textbf{Best}($\nearrow$)         & \textbf{Worst}($\searrow$)       & \textbf{Score}($\nearrow$)         \\ \midrule
HierGNN-BART        & \textbf{0.40}              & 0.20              &   \textbf{0.20}       \\
BART                & 0.29              & \textbf{0.15}              &  0.14        \\
T5-Large            & 0.25              & 0.17              & 0.08         \\
BERTSUMABS          & 0.04              & 0.48              &    *-0.44      \\ \midrule
\textbf{Inf.} & \textbf{Best}($\nearrow$)         & \textbf{Worst}($\searrow$)       & \textbf{Score}($\nearrow$)          \\ \midrule
HierGNN-BART            & 0.35 & \textbf{0.16}         &        0.19  \\
BART                    & \textbf{0.43}             & 0.19              &  \textbf{0.24}        \\
T5-Large                & 0.17              & 0.27              & -0.09         \\
BERTSUMABS              & 0.05              & 0.39              & *-0.34         \\ \midrule
\textbf{Red.}             & \textbf{Best}($\nearrow$)           &           \textbf{Worst}($\searrow$)       & \textbf{Score}($\nearrow$)         \\ \midrule
HierGNN-BART        & \textbf{0.31}              & \textbf{0.21}              &  \textbf{0.10}        \\
BART                & 0.21              & 0.25              &         -0.04 \\
T5-Large            & \textbf{0.31}              & 0.25              &      0.06    \\
BERTSUMABS          & 0.17              & 0.28              &         -0.11 \\ 
\bottomrule
\end{tabular}}
\caption{Detailed summary for the human evaluation in terms of Relevance (Rel.), Informativeness (Inf.) and Redundancy (Red.). We show the proportion of each option to be selected as the Best/Worst among the four candidates. ($\nearrow$) and ($\searrow$) indicates the larger is better and lower is better, respectively. *: HierGNN-BART's scores are significantly (by pair-wise t-test with $p < 0.05$, corrected using Benjamini–Hochberg method to control the False Discovery Rate \cite{benjamini1995controllingfalsedicoveryrate(fdr)} for multiple comparison) better than the corresponding system.}
\label{tab:detalied_human}
\end{table}

\section{Qualitative Case for Graph-Selection Attention}\label{sec:gsa_analysis}

To demonstrate the effectiveness of the graph-selection attention (GSA) on HierGNN, we visualize the graph-selection attention and compare the token attentions whether graph-selection attention is used (See Figure~\ref{fig:gsa_case_study}). It turns out graph-selection attention mostly focuses on the top sentences but still captures the critical information in the latter. In this case, graph-selection attention successfully captures \textit{fifth title in Miami} and \textit{Andy Murray} from the middle part of the article during decoding (marked in blue). In contrast, the model without graph-selection attention continuously produces content about the event \textit{Novak Djokovic beat John Isner} (marked in red).

\end{document}